\newif\ifAIStats
\AIStatsfalse
\newif\ifarXiv
\arXivfalse
\newif\ifWP
\WPfalse

\arXivtrue

\ifAIStats
\documentclass{article}
\usepackage{amsmath,amsfonts,latexsym,graphicx,proceed2e}
\newcommand{\Extra}[1]{}
\newcommand{\Case}[1]{\uppercase{#1}}
\fi

\ifarXiv
\documentclass{article}
\usepackage{amsmath,amsfonts,latexsym,graphicx}
\newcommand{\Extra}[1]{}
\newcommand{\Case}[1]{#1}
\fi

\ifWP
\documentclass[toc]{gtarticle}
\usepackage{amsmath,amsfonts,latexsym,graphicx,epsfig}
\renewcommand{\Extra}[1]{#1}
\newcommand{\Case}[1]{#1}
\fi

\newlength{\picturewidth}
\ifAIStats
\fi
\ifarXiv
\fi
\ifWP
\fi

\newcommand{\Vladimir}{Vladimir }
\newcommand{\DOT}{.}

\newcommand{\given}{\mathrel{|}}

\newcommand{\bbbr}{\mathbb{R}}		
\newcommand{\bbbc}{\mathbb{C}}		
\newcommand{\III}{\mathbb{I}}		
\newcommand{\bbbp}{\mathbb{P}}		
\newcommand{\bbbe}{\mathbb{E}}		
\newcommand{\K}{\mathcal{K}}		
\newcommand{\FFF}{\mathcal{F}}		

\newcommand{\Prob}{\mathop{\bbbp}\nolimits}
\newcommand{\Expect}{\mathop{\bbbe}\nolimits}
\newcommand{\LP}{\mathop{\underline{\bbbp}}\nolimits}
\newcommand{\UP}{\mathop{\overline{\bbbp}}\nolimits}

\newcommand{\sign}{\mathop{{\rm sign}}\nolimits}

\newtheorem{corollary}{Corollary}
\newtheorem{theorem}{Theorem}
\newenvironment{proof}
  {\trivlist\item[\hskip\labelsep\textbf{Proof}]}
  {\endtrivlist}
\newenvironment{Proof}[1]
  {\trivlist\item[\hskip\labelsep\textbf{Proof #1}]}
  {\endtrivlist}
\newcommand{\boxforqed}{\rule{.3em}{1.5ex}}
\newcommand{\qedtext}{\unskip\nobreak\hfil
  \penalty50\hskip1em\null\nobreak\hfil\boxforqed
  \parfillskip=0pt\finalhyphendemerits=0\endgraf}

\newenvironment{remark*}
  {\trivlist\item[\hskip\labelsep{\bfseries Remark}]\relax}
  {\endtrivlist}

\newlength{\IndentI}
\newlength{\IndentII}
\newlength{\IndentIII}
\newlength{\WidthI}
\newlength{\WidthII}
\newlength{\WidthIII}
\ifAIStats
\title{Defensive Forecasting}

\author{Vladimir Vovk\footnotemark[1]\\
\texttt{vovk{\rm@}cs.rhul.ac.uk}\\
\texttt{http://vovk.net}
\And
Akimichi Takemura\footnotemark[2]\\
\texttt{takemura{\rm@}stat.t.u-tokyo.ac.jp}\\
\texttt{http://www.e.u-tokyo.ac.jp/\~{}takemura}
\AND
Glenn Shafer\footnotemark[3] \footnotemark[1]\\
\texttt{gshafer{\rm@}andromeda.rutgers.edu}\\
\texttt{http://glennshafer.com}}
\fi

\ifarXiv
\title{Defensive Forecasting}

\author{Vladimir Vovk\\
\texttt{vovk{\rm@}cs.rhul.ac.uk}\\
\texttt{http://vovk.net}
\and
Akimichi Takemura\\
\texttt{takemura{\rm@}stat.t.u-tokyo.ac.jp}\\
\texttt{http://www.e.u-tokyo.ac.jp/\~{}takemura}
\and
Glenn Shafer\\
\texttt{gshafer{\rm@}andromeda.rutgers.edu}\\
\texttt{http://glennshafer.com}}
\fi

\ifWP
\title{Defensive Forecasting}

\author{Vladimir Vovk\and Akimichi Takemura\and Glenn Shafer}

\twodatestrue

\fi

\begin{document}
\maketitle

\begin{abstract}
  We consider how to make probability forecasts of binary labels.
  Our main mathematical result is that for any continuous
  gambling strategy used for detecting disagreement between the forecasts
  and the actual labels,
  there exists a forecasting strategy
  whose forecasts are ideal
  as far as this gambling strategy is concerned.
  A forecasting strategy obtained in this way
  from a gambling strategy demonstrating a strong law of large numbers
  is simplified and studied empirically.
  \Extra{This working paper is the full version of a paper
  to be published in the AI \& Statistics 2005 proceedings.}
\end{abstract}

\section{\Case{Introduction}}\label{sec:introduction}

Probability forecasting can be thought of
as a game between two players, Forecaster and Reality:

\medskip

\Extra{\noindent
\textsc{Basic Binary Forecasting Protocol}

\noindent
\textbf{Players:} Reality, Forecaster}

\parshape=4
\IndentI  \WidthI
\IndentII \WidthII
\IndentII \WidthII
\IndentII \WidthII
\noindent
FOR $n=1,2,\dots$:\\
  Reality announces $x_n\in{\bf X}$.\\
  Forecaster announces $p_n\in[0,1]$.\\
  Reality announces $y_n\in\{0,1\}$.

\medskip

\noindent
On each round,
Forecaster predicts Reality's move $y_n$
chosen from the \emph{label space},
always taken to be $\{0,1\}$ in this paper.
His move, the \emph{probability forecast} $p_n$,
can be interpreted as the probability he attaches to the event $y_n=1$.
To help 
Forecaster,
Reality presents him with an \emph{object} $x_n$
at the beginning of the round;
$x_n$ are chosen from an \emph{object space} ${\bf X}$.

Forecaster's goal is to produce $p_n$ that agree with the observed $y_n$.
Various results of probability theory,
in particular limit theorems
(such as the weak and strong laws of large numbers, the law of the iterated logarithm,
and the central limit theorem)
and large-deviation inequalities
(such as Hoeffding's inequality),
describe different aspects of agreement between $p_n$ and $y_n$.
For example, according to the strong law of large numbers,
we expect that
\begin{equation}\label{eq:SLLN}
  \lim_{n\to\infty}
  \frac1n\sum_{i=1}^n(y_i-p_i)
  =
  0.
\end{equation}
Such results will be called \emph{laws of probability}
and the existing body of laws of probability
will be called \emph{classical probability theory}.
\Extra{Historically, laws of probability form the core of probability theory.}

In \S\ref{sec:gambling},
following \cite{shafer/vovk:2001},
we formalize Forecaster's goal by adding a third player, Skeptic,
who is allowed to gamble at the odds given by Forecaster's probabilities.
We state a result from \cite{ville:1939} and \cite{shafer/vovk:2001}
suggesting that Skeptic's gambling strategies
can be used as tests of agreement between $p_n$ and $y_n$
and that all tests of agreement between $p_n$ and $y_n$
can be expressed as Skeptic's gambling strategies.
Therefore,
the forecasting protocol with Skeptic
provides an alternative way of stating laws of probability.

As demonstrated in \cite{shafer/vovk:2001},
many standard proof techniques developed in classical probability theory
can be translated into
\ifarXiv continuous strategies for Skeptic. \fi
In \S\ref{sec:result}
we show that for any continuous strategy ${\cal S}$ for Skeptic
there exists a strategy ${\cal F}$ for Forecaster
such that ${\cal S}$ does not detect any disagreement
between the $y_n$ and the $p_n$ produced
by ${\cal F}$.
This result is a ``meta-theorem''
that allows one to move from laws of probability to forecasting algorithms:
as soon as a law of probability is expressed
as a continuous strategy for Skeptic,
we have a forecasting algorithm
that guarantees that this law will hold;
there are no assumptions about Reality,
who may play adversarially.

Our meta-theorem is of any interest
only if one can find sufficiently interesting laws of probability
(expressed as gambling strategies)
that can serve as its input.
In \S\ref{sec:examples} we apply it
to the important properties of unbiasedness in the large and small
of the forecasts $p_n$
((\ref{eq:SLLN}) is an asymptotic version of the former).
The resulting forecasting strategy is automatically unbiased,
no matter what data $x_1,y_1,x_2,y_2,\dots$ is observed.

In \S\ref{sec:experiments} we simplify the algorithm
obtained in \S\ref{sec:examples}
and demonstrate its performance
on some artificially generated data sets.

\section{\Case{The gambling framework for testing probability forecasts}}\label{sec:gambling}

Skeptic is allowed to bet at the odds defined by Forecaster's probabilities,
and he refutes the probabilities
if he multiplies his capital manyfold.
This is formalized as a perfect-information game
in which Skeptic plays against a team composed of Forecaster and Reality:

\medskip

\noindent
\textsc{Binary Forecasting Game I}

\noindent
\textbf{Players:} Reality, Forecaster, Skeptic

\noindent
\textbf{Protocol:}

\parshape=7
\IndentI   \WidthI
\IndentI   \WidthI
\IndentII  \WidthII
\IndentII  \WidthII
\IndentII  \WidthII
\IndentII  \WidthII
\IndentII  \WidthII
\noindent
$\K_0 := 1$.\\
FOR $n=1,2,\dots$:\\
  Reality announces $x_n\in{\bf X}$.\\
  Forecaster announces $p_n\in[0,1]$.\\
  Skeptic announces $s_n\in\bbbr$.\\
  Reality announces $y_n\in\{0,1\}$.\\
  $\K_n := \K_{n-1} + s_n (y_n-p_n)$.

\noindent
\textbf{Restriction on Skeptic:}
Skeptic must choose the $s_n$
so that his capital is always nonnegative
($\K_n \ge 0$ for all $n$)
no matter how the other players move.

\medskip

\noindent
This is a perfect-information protocol;
the players move in the order indicated,
and each player sees the other player's moves
as they are made.
It specifies
both an initial value for Skeptic's capital ($\K_0 = 1$)
and a lower bound on its subsequent values ($\K_n \ge 0$).

Our interpretation,
which will be called the \emph{testing interpretation},
of Binary Forecasting Game I
is that $\K_n$ measures the degree to which Skeptic
has shown Forecaster to do a bad job of predicting $y_i$, $i=1,\dots,n$.

\subsection{\Case{Validity and universality of the testing interpretation}}

As explained in \cite{shafer/vovk:2001},
the testing interpretation is valid and universal
in an important sense.
Let us assume, for simplicity,
that objects are absent (formally, that $|{\bf X}|=1$).
In the case where Forecaster starts
from a probability measure $P$ on $\{0,1\}^{\infty}$
and obtains his forecasts $p_n\in[0,1]$
as conditional probabilities under $P$ that $y_n=1$
given $y_1,\dots,y_{n-1}$,
we have a standard way of testing $P$ and, therefore, $p_n$:
choose an event $A\subseteq\{0,1\}^{\infty}$
(the \emph{critical region})
with a small $P(A)$ and reject $P$ if $A$ happens.
The testing interpretation satisfies the following two properties:
\begin{description}
\item[Validity]
  Suppose Skeptic's strategy is measurable
  and $p_n$ are obtained from $P$;
  $\K_n$ then form a nonnegative martingale
  w.r.\ to $P$.
  According to Doob's inequality
  \cite{ville:1939,doob:1953},
  for any positive constant $C$,
  $\sup_n\K_n\ge C$ with $P$-probability at most $1/C$.
  (If Forecaster is doing a bad job according to the testing interpretation,
  he is also doing a bad job from the standard point of view.)
\item[Universality]
  According to Ville's theorem
  (\cite{shafer/vovk:2001}, \S8.5),
  for any positive constant $\epsilon$
  and any event $A\subseteq\{0,1\}^{\infty}$
  such that $P(A)<\epsilon$,
  Skeptic has a measurable strategy that ensures
  $\liminf_{n\to\infty}\K_n>1/\epsilon$
  whenever $A$ happens,
  provided $p_n$ are computed from $P$.
  (If Forecaster is doing a bad job according to the standard point of view,
  he is also doing a bad job according to the testing interpretation.)
  In the case $P(A)=0$,
  Skeptic actually has a measurable strategy that ensures
  $\lim_{n\to\infty}\K_n=\infty$ on $A$.
\end{description}
The universality of the gambling scenario
of Binary Forecasting Game I
is its most important advantage over von Mises's gambling scenario
based on subsequence selection;
it was discovered by Ville \cite{ville:1939}.

\ifarXiv\subsection{\Case{Continuity of gambling strategies}}\fi

In \cite{shafer/vovk:2001} we constructed
Skeptic's strategies that made him rich
when the statement of any of several key laws of probability theory
was violated.
The constructions were explicit
and lead to
\ifarXiv continuous \fi gambling strategies.
We conjecture that every natural result of classical probability theory
leads to a \ifarXiv continuous \fi strategy for Skeptic.

\Extra{\begin{remark*}According to \cite{atten/dalen:2002},
Brouwer introduced his continuity principle in 1916.
The following quote from Borel's 1912 paper \cite{borel:1912} is popular
(see, e.g., \cite{vonplato:2001}, p.~64):
``a function cannot be calculable unless it is continuous'';
Borel, however, was interested only in the values that computable functions
take on the computable values of the argument.\end{remark*}}

\section{\Case{Defeating Skeptic}}
\label{sec:result}

In this section we prove the main (albeit very simple) mathematical result
of this paper:
for any continuous strategy for Skeptic
there exists a strategy for Forecaster
that does not allow Skeptic's capital to grow,
regardless of what Reality is doing.
Actually, our result will be even stronger:
we will have Skeptic announce his strategy
for each round
before Forecaster's move on that round
rather than making him announce his full strategy at the beginning of the game,
and we will drop the restriction on Skeptic.
Therefore,
we consider the following perfect-information game
that pits Forecaster against the two other players:

\medskip

\noindent
\textsc{Binary Forecasting Game II}

\noindent
\textbf{Players:} Reality, Forecaster, Skeptic

\noindent
\textbf{Protocol:}

\parshape=6
\IndentI   \WidthI
\IndentI   \WidthI
\IndentII  \WidthII
\IndentII  \WidthII
\IndentII  \WidthII
\IndentII  \WidthII
\noindent
$\K_0 := 1$.\\
FOR $n=1,2,\dots$:\\
  Reality announces $x_n\in{\bf X}$.\\
  Skeptic announces continuous $S_n:[0,1]\to\bbbr$.\\
  Forecaster announces $p_n\in[0,1]$.\\
  Reality announces $y_n\in\{0,1\}$.\\
  $\K_n := \K_{n-1} + S_n(p_n) (y_n-p_n)$.

\medskip

\begin{theorem}\label{thm:main}
  Forecaster has a strategy in Binary Forecasting Game II
  that ensures $\K_0\ge\K_1\ge\K_2\ge\cdots$.
\end{theorem}
\begin{proof}
  Forecaster can use the following strategy to ensure $\K_0\ge\K_1\ge\cdots$:
  \begin{itemize}
  \item
    if the function $S_n(p)$ takes the value 0,
    choose $p_n$ so that $S_n(p_n)=0$;
  \item
    if $S_n$ is always positive,
    take $p_n:=1$;
  \item
    if $S_n$ is always negative,
    take $p_n:=0$.
    \qedtext
  \end{itemize}
\end{proof}

\section{\Case{Examples of gambling strategies}}
\label{sec:examples}

In this section we discuss strategies for Forecaster
obtained by Theorem~\ref{thm:main}
from different strategies for Skeptic;
the former will be called \emph{defensive forecasting strategies}.
There are many results of classical probability theory
that we could use,
but we will concentrate on the simple strategy
described in \cite{shafer/vovk:2001}, p.~69,
for proving the strong law of large numbers.

If $S_n(p)=S_n$ does not depend on $p$,
the strategy from the proof of Theorem~\ref{thm:main}
makes Forecaster choose
$$
  p_n
  :=
  \begin{cases}
    0 & \text{if $S_n<0$}\\
    1 & \text{if $S_n>0$}\\
    0 \text{ or } 1 & \text{if $S_n=0$}.
  \end{cases}
$$

The basic procedure described in \cite{shafer/vovk:2001} (p.~69)
is as follows.
Let $\epsilon\in(0,0.5]$ be a small number
(expressing our tolerance to violations
of the strong law of large numbers).
In Binary Forecasting Game I, Skeptic can ensure that
\begin{equation}\label{eq:goal}
  \sup_n\K_n<\infty
  \enspace\Longrightarrow\enspace
  \limsup_{n\to\infty}
  \frac1n\sum_{i=1}^n(y_i-p_i)
  \le
  \epsilon
\end{equation}
using the strategy $s_n=s_n^{\epsilon}:=\epsilon\K_{n-1}$.
Indeed, since
$$
  \K_n
  =
  \prod_{i=1}^n
  (1+\epsilon(y_i-p_i)),
$$
on the paths where $\K_n$ is bounded
we have
$$
  \prod_{i=1}^n
  (1+\epsilon(y_i-p_i))
  \le
  C,
$$$$
  \sum_{i=1}^n
  \ln(1+\epsilon(y_i-p_i))
  \le
  \ln C,
$$$$
  \epsilon
  \sum_{i=1}^n
  (y_i-p_i)
  -
  \epsilon^2
  \sum_{i=1}^n
  (y_i-p_i)^2
  \le
  \ln C,
$$$$
  \epsilon
  \sum_{i=1}^n
  (y_i-p_i)
  \le
  \ln C
  +
  \epsilon^2 n,
$$$$
  \frac1n
  \sum_{i=1}^n
  (y_i-p_i)
  \le
  \frac{\ln C}{\epsilon n}
  +
  \epsilon
$$
(we have used the fact that $\ln(1+t)\ge t-t^2$
when $|t|\le0.5$).
If Skeptic wants to ensure
\begin{multline*}
  \sup_n\K_n<\infty
  \quad\Longrightarrow
\\[-3mm]
  -\epsilon
  \le
  \liminf_{n\to\infty}
  \frac1n\sum_{i=1}^n(y_i-p_i)
  \le
  \limsup_{n\to\infty}
  \frac1n\sum_{i=1}^n(y_i-p_i)
  \le
  \epsilon,
\end{multline*}
he can use the strategy $s_n:=(s_n^{\epsilon}+s_n^{-\epsilon})/2$,
and if he wants to ensure
\begin{equation}\label{eq:implication}
  \sup_n\K_n<\infty
  \enspace\Longrightarrow\enspace
  \lim_{n\to\infty}
  \frac1n\sum_{i=1}^n(y_i-p_i)
  =
  0,
\end{equation}
he can use a convex mixture of $(s_n^{\epsilon}+s_n^{-\epsilon})/2$
over a sequence of $\epsilon$ converging to zero.
There are also standard ways of strengthening (\ref{eq:implication}) to
$$
  \liminf_{n\to\infty}\K_n<\infty
  \enspace\Longrightarrow\enspace
  \lim_{n\to\infty}
  \frac1n\sum_{i=1}^n(y_i-p_i)
  =
  0;
$$
for details, see \cite{shafer/vovk:2001}.

In the rest of this section we will draw
on the excellent survey \cite{dawid:1986}.
We will see how Forecaster defeats
increasingly sophisticated strategies for Skeptic.

\subsection{\Case{Unbiasedness in the large}}

Following Murphy and Epstein \cite{murphy/epstein:1967},
we say that Forecaster is \emph{unbiased in the large}
if (\ref{eq:SLLN}) holds.
Let us first consider the one-sided relaxed version of this property
\begin{equation}\label{eq:one-sided}
  \limsup_{n\to\infty}
  \frac1n\sum_{i=1}^n(y_i-p_i)
  \le
  \epsilon.
\end{equation}
The strategy for Skeptic described above, $S_n(p):=\epsilon\K_n$,
leads to Forecaster always choosing $p_n:=1$;
(\ref{eq:one-sided}) is then satisfied in a trivial way.

Forecaster's strategy corresponding to the two-sided version
\ifAIStats
  \begin{multline}\label{eq:two-sided}
\fi
\ifarXiv
  \begin{equation}\label{eq:two-sided}
\fi
\ifWP
  \begin{equation}\label{eq:two-sided}
\fi
  -\epsilon
  \le
  \liminf_{n\to\infty}
  \frac1n\sum_{i=1}^n(y_i-p_i)
  \le
  \limsup_{n\to\infty}
  \frac1n\sum_{i=1}^n(y_i-p_i)
  \le
  \epsilon
\ifAIStats
  \end{multline}
\fi
\ifarXiv
  \end{equation}
\fi
\ifWP
  \end{equation}
\fi
is not much more reasonable.
Indeed, it can be represented as follows.
The initial capital $1$ is split evenly between two accounts,
and Skeptic gambles with the two accounts separately.
If at the outset of round $n$
the capital on the first account is $\K^1_{n-1}$
and the capital on the second account is $\K^2_{n-1}$,
Skeptic plays $s^1_n:=\epsilon\K^1_{n-1}$ with the first account
and $s^2_n:=-\epsilon\K^2_{n-1}$ with the second account;
his total move is
\ifAIStats
  \begin{multline*}
\fi
\ifarXiv
  \begin{equation*}
\fi
\ifWP
  \begin{equation*}
\fi
  S_n(p)
  :=
  \epsilon\K^1_{n-1}
  -
  \epsilon\K^2_{n-1}
  =
  \epsilon
  \left(
    \prod_{i=1}^{n-1}
    (1+\epsilon(y_i-p_i))
    -
    \prod_{i=1}^{n-1}
    (1+\epsilon(p_i-y_i))
  \right).
\ifAIStats
  \end{multline*}
\fi
\ifarXiv
  \end{equation*}
\fi
\ifWP
  \end{equation*}
\fi
Therefore, Forecaster's move is $p_n:=1$ if
$$
  \sum_{i=1}^{n-1}
  \ln(1+\epsilon(y_i-p_i))
  >
  \sum_{i=1}^{n-1}
  \ln(1+\epsilon(p_i-y_i)),
$$
$p_n:=0$ if
$$
  \sum_{i=1}^{n-1}
  \ln(1+\epsilon(y_i-p_i))
  <
  \sum_{i=1}^{n-1}
  \ln(1+\epsilon(p_i-y_i)),
$$
and $p_n$ can be chosen arbitrarily in the case of equality.
The limiting form of this strategy as $\epsilon\to0$ is:
Forecaster's move is $p_n:=1$ if
$$
  \sum_{i=1}^{n-1}
  (y_i-p_i)
  >
  0,
$$
$p_n:=0$ if
$$
  \sum_{i=1}^{n-1}
  (y_i-p_i)
  <
  0,
$$
and $p_n$ can be chosen arbitrarily in the case of equality.

We can see that unbiasedness in the large does not lead to interesting forecasts:
Forecaster fulfils his task too well.
In the one-sided case (\ref{eq:one-sided}),
he always chooses $p_n:=1$ making
$$
  \sum_{i=1}^n
  (y_i-p_i)
$$
as small as possible.
In the two-sided case (\ref{eq:two-sided}) with $\epsilon\to0$,
he manages to guarantee that
\begin{equation}\label{eq:TooGood1}
  \left|
    \sum_{i=1}^n
    (y_i-p_i)
  \right|
  \le
  1.
\end{equation}
His goals are achieved with categorical forecasts, $p_n\in\{0,1\}$.

In the rest of this section we consider
the more interesting case where $S_n(p)$ depends on $p$.

\subsection{\Case{Unbiasedness in the small}}
\label{subsec:small}

We now consider a subtler requirement that forecasts should satisfy,
which we introduce informally.
We say that the forecasts $p_n$ are \emph{unbiased in the small}
(or reliable, or valid, or well calibrated)
if, for any $p^*\in[0,1]$,
\begin{equation}\label{eq:p}
  \frac
  {\sum_{i=1,\dots,n:p_i\approx p^*} y_i}
  {\sum_{i=1,\dots,n:p_i\approx p^*} 1}
  \approx
  p^*
\end{equation}
provided $\sum_{i=1,\dots,n:p_i\approx p^*} 1$ is not too small.

Let us first consider just one value for $p^*$.
Instead of the ``crisp'' point $p^*$ we will consider a ``fuzzy point''
$I:[0,1]\to[0,1]$ such that $I(p^*)=1$
and $I(p)=0$ for all $p$ outside a small neighborhood of $p^*$.
A standard choice would be something like $I:=\III_{[p_{-},p_{+}]}$,
where $[p_{-},p_{+}]$ is a short interval containing $p^*$
and $\III_{[p_{-},p_{+}]}$ is its indicator function,
but we will want $I$ to be continuous
(it can, however,
be arbitrarily close to $\III_{[p_{-},p_{+}]}$).

The strategy for Skeptic ensuring (\ref{eq:goal})
can be modified as follows.
Let $\epsilon\in(0,0.5]$ be again a small number.
Now we consider the strategy $S_n(p)=S_n^{\epsilon,I}(p):=\epsilon I(p)\K_{n-1}$.
Since
$$
  \K_n
  =
  \prod_{i=1}^n
  (1+\epsilon I(p_i)(y_i-p_i)),
$$
on the paths where $\K_n$ is bounded we have
$$
  \prod_{i=1}^n
  (1+\epsilon I(p_i)(y_i-p_i))
  \le
  C,
$$$$
  \sum_{i=1}^n
  \ln(1+\epsilon I(p_i)(y_i-p_i))
  \le
  \ln C,
$$$$
  \epsilon
  \sum_{i=1}^n
  I(p_i)(y_i-p_i)
  -
  \epsilon^2
  \sum_{i=1}^n
  I^2(p_i)
  (y_i-p_i)^2
  \le
  \ln C,
$$$$
  \epsilon
  \sum_{i=1}^n
  I(p_i)(y_i-p_i)
  -
  \epsilon^2
  \sum_{i=1}^n
  I(p_i)
  \le
  \ln C
$$
(the last step involves replacing $I^2(p_i)$ with $I(p_i)$;
the loss of precision is not great if $I$ is close to $\III_{[p_{-},p_{+}]}$),
$$
  \epsilon
  \sum_{i=1}^n
  I(p_i)(y_i-p_i)
  \le
  \ln C
  +
  \epsilon^2
  \sum_{i=1}^n
  I(p_i),
$$$$
  \frac{\sum_{i=1}^n I(p_i)(y_i-p_i)}{\sum_{i=1}^n I(p_i)}
  \le
  \frac{\ln C}{\epsilon \sum_{i=1}^n I(p_i)}
  +
  \epsilon.
$$
The last inequality shows that the mean of $y_i$ for $p_i$ close to $p^*$
is close to $p^*$
provided we have observed sufficiently many such $p_i$;
its interpretation is especially simple
when $I$ is close to $\III_{[p_{-},p_{+}]}$.

In general, we may consider a mixture of
$S_n^{\epsilon,I}(p)$ and $S_n^{-\epsilon,I}(p)$
for different values of $\epsilon$
and for different $I$ covering all $p^*\in[0,1]$.
If we make sure that the mixture is continuous
(which is always the case for continuous $I$ and finitely many $\epsilon$ and $I$),
Theorem~\ref{thm:main} provides us with forecasts
that are unbiased in the small.

\subsection{\Case{Using the objects}}
\label{subsec:objects}

Unbiasedness, even in the small,
is only a necessary but far from sufficient condition
for good forecasts:
for example, a forecaster who ignores the objects $x_n$
can be perfectly calibrated,
no matter how much useful information $x_n$ contain.
(Cf.\ the discussion of resolution in \cite{dawid:1986};
we prefer not to use the term ``resolution'',
which is too closely connected
with the very special way of probability forecasting
based on sorting and labeling.)
It is easy to make the algorithm of the previous subsection
take the objects into account:
we can allow the test functions $I$ to depend not only on $p$
but also on the current object $x_n$;
$S_n(p)$ then becomes a mixture of
$$
  S_n^{\epsilon,I}(p)
  :=
  \epsilon I(p,x_n)
  \prod_{i=1}^{n-1}
  (1+\epsilon I(p_i,x_i)(y_i-p_i))
$$
and $S_n^{-\epsilon,I}(p)$ (defined analogously)
over $\epsilon$ and $I$.

\Extra{\begin{remark*}Another popular example is:
if the observed sequence of labels is $1,0,1,0,\dots$,
the sequences of forecasts $0.5,0.5,\dots$ and $1,0,1,0,\dots$
are both unbiased in the small,
but the first of them is not as good as the second.
For this example to be covered by our discussion,
the labels $y_n=n\mod2$ should be complemented by objects,
$x_n:=n$.\end{remark*}}

\subsection{\Case{Relation to a standard counter-example}}

Suppose, for simplicity, that objects are absent ($|{\bf X}|=1$).
The standard construction from Dawid \cite{dawid:1985JASA}
showing that no forecasting strategy
produces forecasts $p_n$ that are unbiased in the small for all sequences
is as follows.
Define an infinite sequence $y_1,y_2,\dots$ recursively by
$$
  y_n
  :=
  \begin{cases}
    1 & \text{if $p_n<0.5$}\\
    0 & \text{otherwise},
  \end{cases}
$$
where $p_n$ is the forecast
produced by the forecasting strategy
after seeing $y_1,\dots,y_{n-1}$.
For the forecasts $p_n<0.5$ we always have $y_n=1$
and for the forecasts $p_n\ge0.5$ we always have $y_n=0$;
obviously, we do not have unbiasedness in the small.

Let us see what Dawid's construction gives
when applied to the defensive forecasting strategy
constructed from the mixture of $S_n^{\epsilon,I}(p)$ and $S_n^{-\epsilon,I}(p)$,
as described above,
over different $\epsilon$ 
and different $I$;
we will assume not only that the test functions $I$ cover all $[0,1]$
but also that each point $p\in[0,1]$
is covered by arbitrarily narrow
(concentrated in a small neighborhood of $p$)
test functions.
It is clear that we will inevitably have $p_n\to0.5$
if $p_n$ are produced by the defensive forecasting strategy
and $y_n$ are produced by Dawid's construction.
On the other hand, since all test functions $I$ are continuous
and so cannot sharply distinguish between the cases $p_n<0.5$ and $p_n\ge0.5$,
we do not have any contradiction:
neither the test functions nor any observer
who can only measure the $p_n$ with a finite precision
can detect the lack of unbiasedness in the small.

In this paper we are only interested in unbiasedness in the small
when the test functions $I$ are required to be continuous.
Dawid's construction shows that unbiasedness in the small
is impossible to achieve
if $I$ are allowed to be indicator functions of intervals
(such as $[0,0.5)$ and $[0.5,1]$).
To achieve unbiasedness in the small in this stronger sense,
randomization appears necessary
(see, e.g., \cite{GTP7}).
It is interesting that already a little bit of randomization suffices,
as explained in \cite{kakade/foster:2004}.

\section{\Case{Simplified algorithm}}
\label{sec:experiments}

Let us assume first that objects are absent, $|{\bf X}|=1$.
It was observed empirically
that the performance of defensive forecasting strategies
with a fixed $\epsilon$ does not depend on $\epsilon$ much
(provided it is not too large;
e.g., in the above calculations we assumed $\epsilon\le0.5$).
This suggests letting $\epsilon\to0$
(in particular, we will assume that $\epsilon\ll n^{-2}$).
As the test functions $I$ we will take Gaussian bells $I_j$
with standard deviation $\sigma>0$
located densely and uniformly in the interval $[0,1]$.
Letting $\approx$ stand for approximate equality
and using the shorthand $\sum_{\pm}f(\pm):=f(+)+f(-)$,
we obtain:
$$
  S_n(p)
  =
  \sum_{\pm}
  \sum_j
  (\pm\epsilon)
  I_j(p)
  \prod_{i=1}^{n-1}
  (1\pm\epsilon I_j(p_i)(y_i-p_i))
$$$$
  {}=
  \sum_{\pm}
  \sum_j
  (\pm\epsilon)
  I_j(p)
  \exp
  \left(
    \sum_{i=1}^{n-1}
    \ln(1\pm\epsilon I_j(p_i)(y_i-p_i))
  \right)
$$$$
  {}\approx
  \sum_{\pm}
  \sum_j
  (\pm\epsilon)
  I_j(p)
  \exp
  \left(
    \pm\epsilon
    \sum_{i=1}^{n-1}
    I_j(p_i)(y_i-p_i)
  \right)
$$$$
  {}\approx
  \sum_{\pm}
  \sum_j
  (\pm\epsilon)
  I_j(p)
  \left(
    1\pm\epsilon
    \sum_{i=1}^{n-1}
    I_j(p_i)(y_i-p_i)
  \right)
$$$$
  {}=
  \sum_{\pm}
  \sum_j
  (\pm\epsilon)
  I_j(p)
  \left(
    \pm\epsilon
    \sum_{i=1}^{n-1}
    I_j(p_i)(y_i-p_i)
  \right)
$$$$
  {}\propto
  \sum_j
  I_j(p)
  \sum_{i=1}^{n-1}
  I_j(p_i)(y_i-p_i)
$$
\begin{equation}\label{eq:expression}
  {}=
  \sum_{i=1}^{n-1}
  K(p,p_i)(y_i-p_i),
\end{equation}
where $K(p,p_i)$ is the Mercer kernel
$$
  K(p,p_i)
  :=
  \sum_j
  I_j(p)
  I_j(p_i).
$$
This Mercer kernel can be approximated by
$$
  \int_0^1
  \frac{1}{\sqrt{2\pi}\sigma}
  \exp
  \left(
    -\frac{(t-p)^2}{2\sigma^2}
  \right)
  \frac{1}{\sqrt{2\pi}\sigma}
  \exp
  \left(
    -\frac{(t-p_i)^2}{2\sigma^2}
  \right)
  dt
$$
\begin{align*}
  &\propto
  \int_0^1
  \exp
  \left(
    -\frac{(t-p)^2+(t-p_i)^2}{2\sigma^2}
  \right)
  dt
\\
  &\approx
  \int_{-\infty}^{\infty}
  \exp
  \left(
    -\frac{(t-p)^2+(t-p_i)^2}{2\sigma^2}
  \right)
  dt.
\end{align*}
As a function of $p$,
the last expression is proportional to the density
of the sum of two Gaussian random variables of variance $\sigma^2$;
therefore, it is proportional to
$$
  \exp
  \left(
    -\frac{(p-p_i)^2}{4\sigma^2}
  \right).
$$

To get an idea of the properties of this forecasting strategy,
which we call the \emph{K29} strategy (or algorithm),
we run it and the Laplace forecasting strategy
($p_n:=(k+1)/(n+1)$, where $k$ is the number of 1s observed so far)
on a randomly generated bit sequence of length 1000
(with the probability of $1$ equal to $0.5$).
A zero point $p_n$ of $S_n$ was found using the simple bisection procedure
(see, e.g., \cite{press/etal:1992}, \S\S9.2--9.4, for more sophisticated methods):
(a) start with the interval $[0,1]$;
(b) let $p$ be the mid-point of the current interval;
(c) if $S_n(p)>0$, remove the left half of the current interval;
    otherwise, remove its right half;
(d) go to (b).
We did 10 iterations,
after which the mid-point of the remaining interval was output as $p_n$.
Notice that the values $S_n(0)$ and $S_n(1)$
did not have to be tested.
Our program was written in MATLAB, Version 7,
and the initial state of the random number generator was set to 0.

\begin{figure}[bt]
  \centering
    \includegraphics[width=\picturewidth]{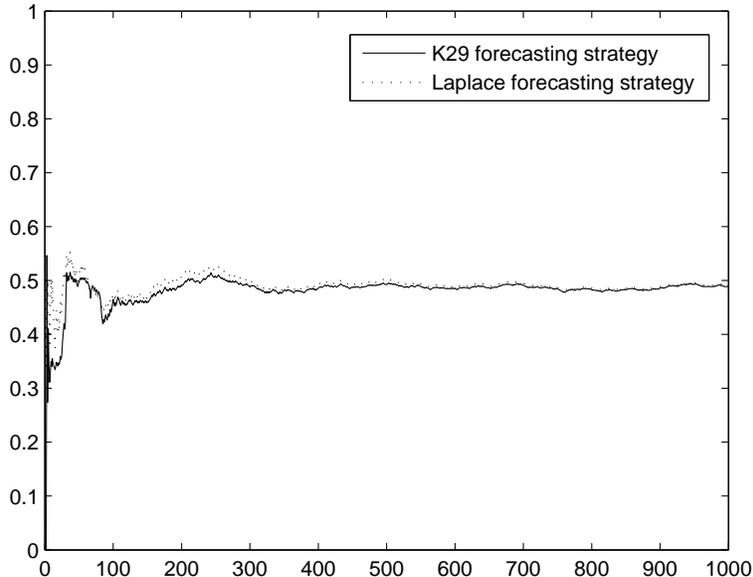}
  \caption{The First $1000$ Probabilities Output
    by the K29 ($\sigma=0.01$) and Laplace Forecasting Strategies
    on a Randomly Generated Bit Sequence
  \label{fig:1000}}
\end{figure}

\begin{figure}[bt]
  \centering
    \includegraphics[width=\picturewidth]{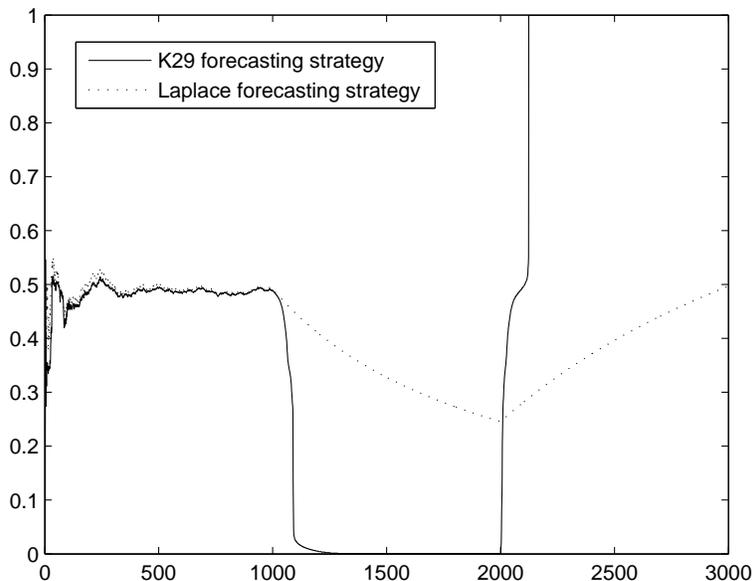}
  \caption{The Probabilities Output
    by the K29 ($\sigma=0.01$) and Laplace Forecasting Strategies
    on a Randomly Generated Sequence of $1000$ Bits
    Followed by $1000$ 0s and $1000$ 1s
  \label{fig:3000}}
\end{figure}

Figure~\ref{fig:1000} shows that the probabilities output
by the K29 ($\sigma=0.01$) and Laplace forecasting strategies
are almost indistinguishable.
To see that these two forecasting strategies can behave very differently,
we complemented the $1000$ bits generated as described above
with $1000$ 0s followed by $1000$ 1s.
The result is shown in Figure~\ref{fig:3000}.
The K29 strategy detects that the probability $p$ of $1$ changes
after the $1000$th round,
and fairly quickly moves down.
When the probability changes again after the $2000$th round,
K29 starts moving toward $p=1$,
but interestingly,
hesitates around the line $p=0.5$,
as if expecting the process to reverse
to the original probability of 1.

The Mercer kernel
$$
  K(p,p_i)
  =
  \exp
  \left(
    -\frac{(p-p_i)^2}{4\sigma^2}
  \right)
$$
used in these experiments is known in machine learning
as the Gaussian kernel
(in the usual parameterization $4\sigma^2$ is replaced by $2\sigma^2$ or $c$);
however, many other Mercer kernels also give reasonable results.

If we start from test functions $I$ depending on the object,
instead of (\ref{eq:expression})
we will arrive at the expression
\begin{equation}\label{eq:S}
  S_n(p)
  =
  \sum_{i=1}^{n-1}
  K((p,x_n),(p_i,x_i))(y_i-p_i),
\end{equation}
where $K$ is a Mercer kernel on the squared product $([0,1]\times{\bf X})^2$.
There are standard ways of constructing such Mercer kernels
from Mercer kernels on $[0,1]^2$ and ${\bf X}^2$
(see, e.g., the description of tensor products and direct sums
in \cite{vapnik:1998,scholkopf/smola:2002}).
For $S_n$ to be continuous,
we have to require that $K$ be \emph{forecast-continuous}
in the following sense:
for all $x\in{\bf X}$ and all $(p',x')\in[0,1]\times{\bf X}$,
$K((p,x),(p',x'))$ is continuous as a function of $p$.
The overall procedure can be summarized as follows.

\medskip

\noindent
\textsc{K29 Algorithm}

\noindent
\textbf{Parameter:} forecast-continuous Mercer kernel $K$ on $([0,1]\times{\bf X})^2$

\parshape=6
\IndentI   \WidthI
\IndentII  \WidthII
\IndentII  \WidthII
\IndentII  \WidthII
\IndentII  \WidthII
\IndentII  \WidthII
\noindent
FOR $n=1,2,\dots$:\\
  Read $x_n\in{\bf X}$.\\
  Define $S_n(p)$ as per (\ref{eq:S}).\\
  Output any root $p$ of $S_n(p)=0$ as $p_n$;\\
  if there are no roots, $p_n:=(1+\sign(S_n))/2$.\\
  Read $y_n\in\{0,1\}$.

\medskip

\noindent
Computer experiments reported in \cite{GTP9}
show that the K29 algorithm
performs well on a standard benchmark data set.
For a theoretical discussion of the K29 algorithm,
\ifarXiv see \cite{GTP8local} (Appendix) \fi
and \cite{GTP11}.

\section{\Case{Related work and directions of further research}}
\label{sec:conclusion}

This paper's methods connect two areas
that have been developing independently so far:
probability forecasting and classical probability theory.
It appears that, when properly developed,
these methods can benefit both areas:
\begin{itemize}
\item
  the powerful machinery of classical probability theory
  can be used for probability forecasting;
\item
  practical problems of probability forecasting
  may suggest new laws of probability.
\end{itemize}

Classical probability theory started from Bernoulli's weak law of large numbers (1713)
and is the subject of countless monographs and textbooks.
The original statements of most of its results
were for independent random variables,
but they were later extended to the martingale framework;
the latter was reduced to its game-theoretic core in \cite{shafer/vovk:2001}.
The proof of the strong law of large numbers used in this paper
was extracted from Ville's \cite{ville:1939} martingale proof
of the law of the iterated logarithm (upper half).

The theory of probability forecasting was a topic of intensive research in meteorology
in the 1960s and 1970s;
this research is summarized in \cite{dawid:1986}.
Machine learning is still mainly concerned with categorical prediction,
but the situation appears to be changing.
Probability forecasting using Bayesian networks
is a mature field;
the literature devoted to probability forecasting using decision trees
and to calibrating other algorithms
is also fairly rich.
So far, however, the field of probability forecasting
has been developing without any explicit connections
with classical probability theory.

Defensive forecasting is indirectly related,
in a sense dual,
to prediction with expert advice
(reviewed in \cite{vovk:2001competitive}, \S4)
and its special case, Bayesian prediction.
In prediction with expert advice
one starts with a given loss function
and tries to make predictions that lead to a small loss
as measured by that loss function.
In defensive forecasting,
one starts with a law of probability
and then makes predictions
such that this law of probability is satisfied.
So the choice of the law of probability
when designing the forecasting strategy
plays a role analogous to the choice of the loss function
in prediction with expert advice.

In prediction with expert advice
one combines a pool of potentially promising forecasting strategies
to obtain a forecasting strategy
that performs not much worse than the best strategies in the pool.
In defensive forecasting
one combines strategies for Skeptic
(such as the strategies corresponding to different test functions $I$
and different $\pm\epsilon$ in \S\ref{sec:examples})
to obtain one strategy achieving an interesting goal
(such as unbiasedness in the small);
a strategy for Forecaster
is then obtained using Theorem~\ref{thm:main}.
The possibility of mixing strategies for Skeptic
is as fundamental in defensive forecasting
as the possibility of mixing strategies for Forecaster
in prediction with expert advice.

This paper continues the work
started by Foster and Vohra~\cite{foster/vohra:1998}
and later developed in, e.g., \cite{lehrer:2001,sandroni/etal:2003,GTP7}
(the last paper replaces the von Mises--style framework
of the previous papers with a martingale framework,
as in this paper).
The approach of this paper is similar to that
of the recent paper \cite{kakade/foster:2004},
which also considers deterministic forecasting strategies
and continuous test functions for unbiasedness in the small.

The main difference of this paper's approach
from 
the bulk of work in learning theory
is that we do not make any assumptions about Reality's strategy.

The following directions of further research
appear to us most important:
\begin{itemize}
\item
  extending Theorem~\ref{thm:main} to other forecasting protocols
  (such as multi-label classification)
  and designing efficient algorithms for finding the corresponding $p_n$;
\item
  exploring forecasting strategies corresponding to:
  (a) Hoeffding's inequality, (b) the central limit theorem,
  (c) the law of the iterated logarithm
  (all we did in this paper was to slightly extend the strong law of large numbers
  and then use it for probability forecasting).
\end{itemize}

\subsection*{Acknowledgments}

We are grateful to the participants of the PASCAL workshop
``Notions of complexity: information-theoretic, computational and statistical approaches''
(October 2004, EURANDOM)
who commented on this work
and to the anonymous referees for useful suggestions.
This work was partially supported
by BBSRC (grant 111/BIO14428),
EPSRC (grant GR/R46670/01),
MRC (grant S505/65),
Royal Society,
and, especially,
the Superrobust Computation Project
(Graduate School of Information Science and Technology,
University of Tokyo).
\ifarXiv

\fi
\end{document}